\documentclass[11pt,a4paper]{article}
\usepackage[utf8]{inputenc}
\usepackage{natbib}                        
\PassOptionsToPackage{breaklinks}{hyperref} 
\usepackage[hyperref]{acl2021}
\usepackage{times}
\usepackage{latexsym}

\usepackage{soul}
\usepackage{graphicx}
\usepackage{amsmath}
\usepackage[mathscr]{euscript} 
 \let\mathscr\relax
\usepackage[ruled,vlined]{algorithm2e}
\usepackage[scr]{rsfso}
\usepackage{subcaption}
\usepackage{stackengine}
\usepackage{lscape}
\usepackage{adjustbox}
\usepackage{bm}
\usepackage{multirow}

\usepackage{array}

\usepackage{multicol,lipsum}
\usepackage{lipsum}
\usepackage{mathtools}
\usepackage{cuted}
\usepackage{microtype}
\aclfinalcopy 

\title{Textual Data Distributions: Kullback Leibler Textual Distributions Contrasts on GPT-2 Generated Texts, with Supervised, Unsupervised Learning on Vaccine \& Market Topics \& Sentiment}
\aclfinaltrue

\author{ Jim Samuel \\
  \textit{jim@aiknowledgecenter.com}\\
  University of Charleston \\\And
  Ratnakar Palle \\
  \textit{rrpalle@gmail.com}\\
  Apple Inc.\\\And
  Eduardo Correa Soares \\
  \textit{ecsoares@yahoo.com}\\
  Cerence B. V.\\}

\date{May, 2021}

\begin{document}

\maketitle

\begin{abstract}
Efficient textual data distributions (TDD) alignment and generation are open research problems in textual analytics and NLP. It is presently difficult to parsimoniously and methodologically confirm that two or more natural language datasets belong to similar distributions, and to identify the extent to which textual data possess alignment. This study focuses on addressing a segment of the broader problem described above by applying multiple supervised and unsupervised machine learning (ML) methods to explore the behaviour of TDD by (i) topical alignment, and (ii) by sentiment alignment. Furthermore we use multiple text generation methods including fine-tuned GPT-2, to generate text by topic and by sentiment. Finally we develop a unique process driven variation of Kullback-Leibler divergence (KLD) application to TDD, named \textit{``KL Textual Distributions Contrasts''} \texttt(KL-TDC) to identify the alignment of machine generated textual corpora with naturally occurring textual corpora. This study thus identifies a unique approach for generating and validating TDD by topic and sentiment, which can be used to help address sparse data problems and other research, practice and classroom situations in need of artificially generated topic or sentiment aligned textual data.    
\end{abstract}

\subsection*{Keywords:} \textit{Textual data distributions, supervised learning, unsupervised learning, Kullback-Leibler divergence, sentiment, textual analytics, text generation, vaccine, stock market}

\section{Introduction}


Recent developments in natural language processing (NLP) have shown that the state of the art in many common tasks is highly dependent on models with a larger number of parameters trained on colossal amounts of data \citep{devlin2018bert,  radford2019language, brown2019verbnet}. While the advances in computing power and technologies allow researchers and developers to increase the number of parameters in their models, attempts to increase the size of datasets reveal many challenges that are hard to overcome. 
There is a need to develop capabilities to align and generate textual data distributions (TDD) by topic and by other parameters such as sentiment. 
Just as the use of quantitative distributions have enabled much scientific progress across disciplines, and 
so also TDD generation capabilities would be immensely useful for the advancement of research in textual analytics and NLP. Such machine generated TDD can be extremely useful in the development and testing of new methods and technologies, and can also be a valuable tool in classrooms - it can be used widely in curricula and for workforce training purposes. These capabilities could be used in a wide range of applications as well, such as for augmenting behavioral finance by generating text aligned with the distribution of ``seed" posts on social media which could be used to identify current and impending target group behavior.    

Additionally, there are a number of languages that are not as representative on the internet as English is, because they are not spoken by as many people or because of the lack of economic power of linguistic groups. This highlights the importance of having efficient textual distributions generation methods which can be extended to other languages as well.  
Finally, even for the English language, in textual data-rich domains, restrictions concerning the source of the data may reduce the availability of samples in areas such as medicine, for instance. A number of techniques have been proposed to increase the amount of textual data, from simple heuristics to complex neural networks. However, a fundamental problem remains understudied: how do we test and ensure that the distributions of the artificially generated data are aligned with those of the real world data of interest? In this paper, we use topic classification and sentiment analysis on Twitter datasets, generate textual data, and identify metrics to test TDD.

In this study, we employ tweets from `Vaccine' and `Market' keywords filtered Twitter data, and use the preprocessed tweets text data as input. We have three levels of outputs: first, we test supervised machine learning (ML) methods with and without keywords, and review classification accuracy, second we test unsupervised ML methods and third we generate text using three different ML methods to test for alignment of distributions using an adapted form of the Kullback Leibler Divergence (KLD) test \citep{kullback1951information}.  

We use a priori knowledge of the topics, the sentiment and the distributions. Our conceptual measure of success will therefore be the degree to which algorithms are able to learn and generate text with similar distributions, based on classified data and known distributions from our preprocessed and organized original data sets. To the best of our knowledge, there is no widely accepted method to test whether two or more data sets of language data, natural and machine-generated are aligned with respect to their distributions and topic or sentiment coverage. There are some useful but weakly related studies in recent publications which we have mentioned in our literature review below. However, we were not able to find a general approach or solution to this problem, which could be straightforwardly adopted and applied. Therefore, our overarching purpose is to propose and test such an approach for generating and testing alignment of textual distributions. 

\section{Literature Review}
Given our interest in topic classification and sentiment analysis based on TDD and text generation using multiple machine learning methods, our literature review falls broadly into a few key categories: a) Past research that addresses textual analytics and topic identification, b) machine learning methods for textual data and NLP, c) statistical methods for TDD and d) text generation and data augmentation. 
Illustratively, a recent work on logical natural language generation (NLG) provides us with interesting input on logic in natural language understanding \citep{chen2020logical}. They identify the weaknesses in current NLP and NLG strategies which primarily depend on “surface-level” pairing and links between words and phrases, which is useful for some NLP tasks, such as association mining. However, such surface level methods are unable to go into the depth of the text to make sense of the textual artifacts and draw logical inferences, which maybe could point towards an approach for TDD and topic alignment. This remains an open problem in NLP and NLG, and the clear articulation of the problem, as well as the strategy highlighted by Chen et al. to address these issues is insightful \citep{chen2020logical}.  

\subsection{Overview of Methods for NLP tasks and text-to-number approaches}
One of the major and early stage decisions for textual analytics and NLP projects involves the selection of suitable quantitative representations for text corpora. A broad range of strategies and methods exist, depending on the purpose, the context and the nature of text corpora. Madureira and Schlangen provide a valuable summary of state of the art textual states representation, with a focus on reinforcement learning, covering extant methods across a range of machine learning, deep learning and neural network approaches \citep{madureira2020overview}. They highlight the absence of agreement, in spite of reasonable common ground, for the textual states representation problem and we see this as arising out of the need for a dominant generic solution, which could universally cater to multiple NLP goals. SzymańSki compares text representation methods contextualized to ``knowledge representation” for  ``for documents categorization"  \citep{szymanski2014comparative}. The study defines ``Explicit Semantic Analysis'' (ESA) as a hybrid method combining multiple methods that use ``content  and  referential  approaches'' \citep{gabrilovich2007computing}: with the content approach, the representation of text corpora can be driven by a combination of bag of words (BOW) and N-grams which look at intrinsic substance within a textual corpus; with the referential approach, identification  of concepts within a textual corpus is attempted by using similarity measures against a referential set of concepts. The referential set could consist of a very large cluster of concepts such as all Wikipedia articles, or could consist of a relatively narrowed set using heuristics or logical deduction. The study compares the effectiveness of common representation methods: cosine kernel, n-grams (letters), n-grams (words), Explicit Semantic Analysis (ESA), links, higher order references (HOR) and compression. The cosine kernel refers to the use of cosine measures ``between article vectors created using TF-IDF (term frequency - inverse document frequency) weighting''. N-grams identify letters and words sequences by frequency of usage within the text corpus. The compression method for testing for similarity uses a ratio of the size of algorithmically compressed combined textual corpora to the sum of the size of algorithmically compressed individual textual corpora. Links refer to text corpora with direct association, and HOR is “higher order references” – which extend the associations, usually with a reduced weight.


Neural learning methods have been widely used to address NLP challenges successfully. A conceptual basis is provided for the relative success of neural methods against non-neural methods, credited to the observation that \textit{``Non-neural NLP methods usually heavily rely on the discrete handcrafted features''} \citep{qiu2020pre}. In their survey of the usage of pre-trained language models for NLP purposes, \citet{qiu2020pre} also posit that the success of neural methods is often driven by their use of ``low-dimensional and dense vectors'' to better reflect or ``represent the syntactic or semantic features" of textual corpora. 
However, such neural representations are subject to ``specific NLP tasks" and therefore may subscribe to potential overfitting. They also highlighted the effectiveness of BERT (\texttt{Bidirectional Encoder Representations from Transformers}, one of the largest pretrained language models) for sentiment analysis (associating human sentiment score or class to textual corpora) and named entity recognition (NER, disambiguates sentences into entity classes of words). BERT’s effectiveness in addressing general NLP tasks with common textual corpora, as compared to traditional machine learning methods for classification, is well supported \citep{gonzalez2020comparing}. Other surveys and extant research have reviewed NLP tools and industry applications \citep{kalyanathaya2019advances}, NLP attention mechanisms \citep{hu2019introductory}, NLP for opinion classification \citep{othman2015using}, and deep learning contributions to NLP applications, tasks and objectives \citep{torfi2020natural}.

Generative Pre-trained Transformer (GPT) models are deep learning based pretrained autoregressive language models that generate human-like text, and can be fine tuned to adapt to localized contexts. Neural text generation methods have rapidly grown over the past few years, and have yielded rich results, being broadly classified into ``transfer learning" (such as ``Embeddings from Language Models'' - ELMo and BERT), and ``deep contextual language modeling" (such as GPT, GPT-2 and GPT-3) \citep{ji2020amazing}. This study uses a locally fine-tuned model based on GPT-2 to generate text by topics: \texttt{Vaccine and Market}.

\subsection{Data Augmentation \& Distributions}
Most studies on Data Augmentation test only the improvements in accuracy of the classifiers (in general neural learning methods)  on some supervised learning task with and without data augmentation (see, for instance, \cite{hou2018sequence, wei2019eda, guo2019augmenting} and many others). However, testing the distributions is not a common practice in the literature on textual data generation. Notably, there are two recent papers that go beyond testing accuracy of a neural learning method: \textit{“Text data augmentation made simple by leveraging nlp cloud APIs”} \citep{coulombe2018text} and \textit{“Quantifying the Evaluation of Heuristic Methods for Textual Data Augmentation”} \citep{kashefi2020quantifying}. Coulombe's paper summarizes data augmentation techniques for textual data and attempts to evaluate them. The evaluation is formalized in some constraints: ``Rule of respect the statistical distribution", ``Golden Rule of Plausibility", ``Semantic invariance rule" and ``Telephone Game Rule of Thumb". However, the test focuses on accuracy of classifying movie reviews into some categories. No further test on the distributions was carried out, even if they are sketched as an important criterion \citep{coulombe2018text}. In the ``Quantifying the evaluation of Heuristic Methods for Textual Data Augmentation” paper, the main proposal is to use an evaluation approach to multiple heuristics and augmented datasets for classification tasks \citep{kashefi2020quantifying}. The augmented datasets were evaluated in terms of accuracy (whether Recurrent neural networks (RNNs) and Convolutional neural networks (CNNs) were classifying the texts in the right class in a supervised learning task) and in a metric called ``hard to distinguish". This metric was calculated as the Kullback \& Leiber divergence (KLD) \citep{kullback1951information}. KLD is used to calculate how much a probability distribution diverges from another as a measure of information gain if samples of the later were used instead of the former. The smaller this score is, the harder it is to distinguish the two distributions. 

\subsection{Topic, Sentiment: Similarity modeling}
Similarity modeling is another interesting concept which has significant implications for a wide usage in NLP, and has strong relevance to our interest in topical distributions of textual data. \citet{janusz2012unsupervised} develop a similarity model, the primary purpose of which is stated as being for ``semantic information retrieval task or semantic clustering". They discuss and rely on Tversky’s Similarity Model which works well in the context of judgements made by human intelligence \citep{tversky1977features}. They propose ``bireducts'' algorithms  ``which correspond to different contexts or points of view for evaluation of document resemblance'', and combine this algorithmic approach with Tversky’s  equation to posit a novel approach to similarity modeling. In fact, clustering is a promising approach for topic modelling as well as for other NLP tasks. Even though \citet{selosse2020textual} focus on data summarization, they propose a unique co-clustering approach, which may be useful for topic alignment. Their method leads to the identification of ``homogeneous co-clusters", which is also accomplished by a range of alternative algorithms, but the study also adds value by contrasting ``noisy co-clusters" with ``significant co-clusters, which is particularly useful for sparse document-term matrices''.

\citet{garg2021unsupervised} study related concepts of ``Semantic Similarity, Textual Entailment, Expression Diversity and Fluency"  to address the challenges of providing satisfactory heterogeneity of communicative interactions for artificial agents responding to human inquiries. They measure the performance effectiveness of their Reinforcement Learning approach by referencing ``the automated metric as the reward function", which is somewhat of a concern as it appears to pose a self-referential challenge. The automated metric itself is a measure of the ``quality of contextual paraphrases''. It is not clear if the authors had a rationale to address this weakness, nevertheless, the study provides interesting domain insights.

\subsection{Comments on contrast}
It is worth highlighting that most of the literature on artificial textual data generation (mainly data augmentation) uses neural learning methods, which are de facto based on low-dimensional and dense vectors. However, as mentioned in section 2.1, we have found only one paper that explicitly tests the distributions of the data, which is based on KLD generated from word embeddings. As all other papers focus on improvements in accuracy of a set of supervised-learning tasks using neural networks, we took a different approach looking at both supervised learning and unsupervised learning tasks. Namely, beyond using neural learning methods for topic classification, we are also interested in testing an unsupervised learning algorithm: clustering. We expect that unsupervised learning methods will be a less costly way of testing data distributions and topic alignment, which may also be incorporated in other methods.


\section{Propositions \& Methods}
This section outlines the propositions (quasi-hypotheses) and methods for our study: the conceptual intent and expectations, description of data utilized in the study, theoretical basis and metrics used to build and evaluate the models respectively. 
We initiate our process by applying supervised classification methods for topic and sentiment classification, followed by unsupervised text clustering, and text generation with three methods. We select GPT2 fine tuned models for generating the final texts, and use a unique distribution construction process for applying KLD tests to gauge similarity of distributions between the original and generated texts. 


\subsection{Intent and Expectations: Propositions}
Our research is anchored upon: 
\begin{itemize}
\item 1: Conceptual distinctions of TDD on the basis of 
    \begin{itemize}
    \item a) topic (such as named entity or keyword), specifically the topics of vaccine and market are used in this study.  
    \item b) sentiment (such as positive or negative classes or scores, as generated using popular NLP sentiment dictionaries).  
\end{itemize}
We focus on the study of data distributions qualified by 1a and 1b (topic and sentiment) in the current study.
\item 2: Methodological comparison based on the applications of 
    \begin{itemize}
    \item a) Supervised ML classification: Logistic regression, Support Vector Machines (SVM), and Na\"ive Bayes. \item b)  Unsupervised ML classification: a) Hierarchical Agglomerative Clustering (HAC) and b) K-Means Clustering (KMC).
    \item c) Three ML text generation methods: Direct probabilistic, Recurrent Neural Networks ((RNN) and Long-Short-Term-Memory (LSTM), and fine tuned GPT2. 
\end{itemize}
\end{itemize}
The final step is to validate alignment of generated TDD with naturally occurring TDD using adapted KLD as described in the sections below.    

\subsubsection{Propositions: TDD by Topic \& Sentiment}
Based on the above, we worked on addressing key research interests listed below. We developed processes to explore original textual distributions, machine-generate text and evaluate whether generated and original textual datasets are aligned by distribution. We did this based on: \\
\textbf{(i)} text trained by topic category and\\
\textbf{(ii)} text trained by sentiment class.\\
The ``trained by" is applicable where at least one of the datasets is machine generated, and ``based on" refers to comparison of naturally occurring textual data. 
Based on our conceptualization thus far, we hypothesize two propositions, the first being that: 
\begin{itemize}
\item P1: \emph{TDD categorized by topic and sentiment can be contrasted using supervised and unsupervised learning methods}.

 
Additionally, based on our study of distribution identification and alignment methods posited above, we hope to be able to improve the quality of textual data generation by comparing and selecting from a) a direct probabilistic distribution based text generation, b) RNN - LSTM approach, and c) text-generation with fine tuned GPT2 models. 

Direct-probabilistic and RNN-LSTM methods generate textual data with a fair degree of alignment with the input data. However, their vocabulary is limited to the scope of the textual input, and therefore we also use fine tuned GPT2 modesl to generate data. We generate data from topic and sentiment classification labels assigned natural data and explore improving models for generating higher quality data which will be better aligned with topic or sentiment based seed input.

Based on our conceptualization, the second proposition is that: 
\item P2: \emph{It is possible to obtain satisfactory alignment of artificially generated TDD with naturally occurring TDD, by topic and sentiment classifications}.
\end{itemize}
We discuss the measure of success and improvements in the Theory and Metrics sections below.

Presently, as a sub-goal, we intend to heuristically evaluate the semantic quality of generated text by human judgement, supported by textual analytics and data visualization of generated text. We will analyze term and phrase (N-gram) frequencies, alignment with desired topic, and also explore comparisons with commonly known generative pre-trained models. We will also compare and evaluate the results by applying our findings to additional new small random samples from our main data. We mention this as a sub-goal because even if the generated text were in garbled sequences of words and did not make semantic sense, yet it could still serve the overarching purpose of algorithmic textual data distribution alignment.

\subsection{Data} \label{datasection}
We acquired Twitter data on multiple topics, downloaded from Twitter with a developer account API using a broad range of keywords. The present research stream initially focuses on tweets associated with two different topics - \emph{``vaccine''} and \emph{``stock market''} for this study. We initiated our process with two small random samples of two hundred tweets from the each of the two main tweets datasets (Over one million Tweets).  The downloaded data have about ninety variables and we extract only the \emph{``Text''} variable for our analyses and modeling.

\subsubsection{Data subsets} The main data were filtered to create a subset of data  based on the account location by country as \emph{United States}, for each of the topics. Tweets containing urls were deleted to exclude spam, and separately, abusive words were algorithmically replaced with \emph{``abusv123987''} (a unique enough string with an extremely low likelihood of natural occurrence in Tweets). A random sample generation process, without replacement, was applied to subset 200 randomly selected representative tweets for each topic along with a corresponding label ( \emph{M} for market and \emph{V} for vaccine). The two datasets were then joined and randomized in order to create our pilot data of 400 topic-labelled tweets.  

\subsubsection{Data preparation for trial} The sample data were cleaned and processed using standard NLP preprocessing tools in \emph{R} and \emph{Python}. The \emph{Text} variable was extracted, stripped of special characters and cleaned. The \emph{Text} variable was deliberately not stemmed or lemmatized because of our interest in both words and phrases, and in the semantic structure of tweets. In addition to the topic labels (\emph{M} for market and \emph{V} for vaccine) in the 400 tweets dataset, we also created an additional sentiment label. The tweets were assigned a sentiment score using the SentimentR package, and the default Jockers dictionary. All tweets with scores greater than 0, were classified as positive tweets and all tweets with scores less than 0 were classified as negative tweets. Neutral tweets with a sentiment score of 0 were excluded, to create a positive - negative labeled dataset of 342 tweets.

\hfill
\break
We used around 400 tweets for the pilot modeling phase to test our experimental classification concepts, models and code, and about 10,000 tweets for our hierarchical models and code, and then repeated the process, as described above and minus creating data subsets, for the final reported classification analysis with complete datasets. 

\subsection{Theory and Metrics }

As mentioned before, our project aims at studying TDD, and improve textual data generation associated with topic and sentiment alignment. 
Our starting point are baseline supervised and unsupervised models. One of the goals of our approach is to study and develop metric/s to evaluate the fitness of the generated data to improve performance in other tasks. The following metrics will be used to evaluate our models:

a. accuracy, including precision, recall and F1-score on the test set in supervised learning tasks before and after addition of generated data;

b. overall accuracy, including precision, recall and F1-score in the unsupervised tasks before and after addition of generated data;

c. Customized variation of Kullback and Leiber (1951)'s divergence application to evaluate how much two data sets are draw out of the same distribution or not.

We evaluate machine generated text against our originally collected naturally occurring data, using a random sample subset as a baseline for evaluation. 

\small
\begin{table*}
\centering
\begin{tabular}{|cccccccccccccccc} \cline{1-14}
\multicolumn{2}{|c}{Model} & \multicolumn{6}{|c|}{With Keyword} & \multicolumn{6}{c|}{Without Keyword} \\ \cline{1-14}

\multirow{3}{*}{SVM} & \multicolumn{1}{c|}{} & \multicolumn{6}{c|}{ \begin{tabular}[t]{| c | c | c |}
  \firsthline
  \multicolumn{1}{|l|}{} & \multicolumn{1}{l|}{0} & \multicolumn{1}{l|}{1} \\
  \hline
  0 & 584 & 12 \\
  \hline
  1 & 11 & 1780 \\
  \hline
  \end{tabular}  } & \multicolumn{6}{c|}{  \begin{tabular}[t]{| c | c | c |}
  \firsthline
  \multicolumn{1}{|l|}{} & \multicolumn{1}{l|}{0} & \multicolumn{1}{l|}{1} \\
  \hline
  0 & 384 & 167 \\
  \hline
  1 & 134 & 1702 \\
  \hline
  \end{tabular}   } \\ \hline

\multirow{3}{*}{\begin{tabular}{@{}c@{}}Na\"ive Bayes\\(Bernoulli)\end{tabular}} & \multicolumn{1}{c|}{}
& \multicolumn{6}{c|}{  \begin{tabular}[t]{| c | c | c |}
  \firsthline
  \multicolumn{1}{|l|}{} & \multicolumn{1}{l|}{0} & \multicolumn{1}{l|}{1} \\
  \hline
  0 & 594 & 2 \\
  \hline
  1 & 10 & 1781 \\
  \hline
  \end{tabular}  } & \multicolumn{6}{c|}{  \begin{tabular}[t]{| c | c | c |} 
  \firsthline
  \multicolumn{1}{|l|}{} & \multicolumn{1}{l|}{0} & \multicolumn{1}{l|}{1} \\
  \hline
  0 & 468 & 83 \\
  \hline
  1 & 246 & 1590 \\
  \hline
  \end{tabular}    } \\ \hline

\multirow{3}{*}{\begin{tabular}{@{}c@{}}Logistic\\Regression\end{tabular}} & \multicolumn{1}{c|}{}
& \multicolumn{6}{c|}{  \begin{tabular}[t]{| c | c | c |}
  \firsthline
  \multicolumn{1}{|l|}{} & \multicolumn{1}{l|}{0} & \multicolumn{1}{l|}{1} \\
  \hline
 0 & 582 & 14 \\
  \hline
  1 & 6 & 1785 \\
  \hline
  \end{tabular}  } & \multicolumn{6}{c|}{  \begin{tabular}[t]{| c | c | c |}
  \firsthline
  \multicolumn{1}{|l|}{} & \multicolumn{1}{l|}{0} & \multicolumn{1}{l|}{1} \\
  \hline
 0 & 383 & 168 \\
  \hline
  1 & 126 & 1710 \\
  \hline
  \end{tabular}    } \\ \hline

\hline
\end{tabular}
\caption{\small{Confusion Matrices of Supervised Learning Based Classifiers  for Topic Classification.}}
\label{table:TopicConfusionMatrix}
\end{table*} \normalsize


\small
\begin{table*}
\centering
\begin{tabular}{|ccccccccccccccc} \cline{4-15}
\multicolumn{3}{c|}{} & \multicolumn{6}{|c|}{With Keyword} & \multicolumn{6}{c|}{Without Keyword} \\ \cline{1-15}

\multicolumn{1}{|c|}{Model} & \multicolumn{2}{c|}{Class} & \multicolumn{2}{c|}{Precision} & \multicolumn{2}{c|}{Recall} & \multicolumn{2}{c|}{F1-Score} & \multicolumn{2}{c|}{Precision} & \multicolumn{2}{c|}{Recall} & \multicolumn{2}{c|}{F1-Score} \\ \hline

\multirow{2}{*}{SVM} & \multicolumn{2}{|c|}{StM} & \multicolumn{2}{c|}{0.98} & \multicolumn{2}{c|}{0.98} & \multicolumn{2}{c|}{0.98} & \multicolumn{2}{c|}{0.74} & \multicolumn{2}{c|}{0.70} & \multicolumn{2}{c|}{0.72} \\

 & \multicolumn{2}{|c|}{Vac} & \multicolumn{2}{c|}{0.99} & \multicolumn{2}{c|}{0.99} & \multicolumn{2}{c|}{0.99} & \multicolumn{2}{c|}{0.91} & \multicolumn{2}{c|}{0.93} & \multicolumn{2}{c|}{0.92} \\ \cline{1-15}

\multirow{2}{*}{\begin{tabular}{@{}c@{}}Na\"ive Bayes\\(Bernoulli)\end{tabular}} & \multicolumn{2}{|c|}{StM} & \multicolumn{2}{c|}{0.98} & \multicolumn{2}{c|}{1.00} & \multicolumn{2}{c|}{0.99} & \multicolumn{2}{c|}{0.66} & \multicolumn{2}{c|}{0.85} & \multicolumn{2}{c|}{0.74} \\ 

 & \multicolumn{2}{|c|}{Vac} & \multicolumn{2}{c|}{1.00} & \multicolumn{2}{c|}{0.99} & \multicolumn{2}{c|}{1.00} & \multicolumn{2}{c|}{0.95} & \multicolumn{2}{c|}{0.87} & \multicolumn{2}{c|}{0.91} \\ \cline{1-15}

\multirow{2}{*}{\begin{tabular}{@{}c@{}}Logistic\\Regression\end{tabular}} & \multicolumn{2}{|c|}{StM} & \multicolumn{2}{c|}{0.99} & \multicolumn{2}{c|}{0.98} & \multicolumn{2}{c|}{0.98} & \multicolumn{2}{c|}{0.75} & \multicolumn{2}{c|}{0.70} & \multicolumn{2}{c|}{0.72} \\ 

 & \multicolumn{2}{|c|}{Vac} & \multicolumn{2}{c|}{0.99} & \multicolumn{2}{c|}{1.00} & \multicolumn{2}{c|}{0.99} & \multicolumn{2}{c|}{0.91} & \multicolumn{2}{c|}{0.93} & \multicolumn{2}{c|}{0.92} \\ 

\hline
\end{tabular}
\caption{\small{Perfomance of Supervised Learning Based Classifiers for Topic Classification.}}
\label{table:TopicPerformance}
\end{table*} \normalsize

\small
\begin{table*}
\centering
\begin{tabular}{|cccccccccccccc} \cline{1-14}
\multicolumn{2}{|c}{Model} & \multicolumn{6}{|c|}{Unbalanced Dataset} & \multicolumn{6}{c|}{Balanced Dataset} \\ \cline{1-14}

\multirow{3}{*}{SVM} & \multicolumn{1}{c|}{} & \multicolumn{6}{c|}{ \begin{tabular}[t]{| c | c | c |}
  \firsthline
  \multicolumn{1}{|l|}{} & \multicolumn{1}{l|}{0} & \multicolumn{1}{l|}{1} \\
  \hline
  0 & 436 & 346 \\
  \hline
  1 & 292 & 896 \\
  \hline
  \end{tabular}  } & \multicolumn{6}{c|}{  \begin{tabular}[t]{| c | c | c |}
  \firsthline
  \multicolumn{1}{|l|}{} & \multicolumn{1}{l|}{0} & \multicolumn{1}{l|}{1} \\
  \hline
  0 & 192 & 177 \\
  \hline
  1 & 159 & 442 \\
  \hline
  \end{tabular}   } \\ \hline

\multirow{3}{*}{\begin{tabular}{@{}c@{}}Na\"ive Bayes\\(Bernoulli)\end{tabular}} & \multicolumn{1}{c|}{}
& \multicolumn{6}{c|}{  \begin{tabular}[t]{| c | c | c |}
  \firsthline
  \multicolumn{1}{|l|}{} & \multicolumn{1}{l|}{0} & \multicolumn{1}{l|}{1} \\
  \hline
  0 & 594 & 2 \\
  \hline
  1 & 10 & 1781 \\
  \hline
  \end{tabular}  } & \multicolumn{6}{c|}{  \begin{tabular}[t]{| c | c | c |} 
  \firsthline
  \multicolumn{1}{|l|}{} & \multicolumn{1}{l|}{0} & \multicolumn{1}{l|}{1} \\
  \hline
  0 & 468 & 83 \\
  \hline
  1 & 246 & 1590 \\
  \hline
  \end{tabular}    } \\ \hline

\multirow{3}{*}{\begin{tabular}{@{}c@{}}Logistic\\Regression\end{tabular}} & \multicolumn{1}{c|}{}
& \multicolumn{6}{c|}{  \begin{tabular}[t]{| c | c | c |}
  \firsthline
  \multicolumn{1}{|l|}{} & \multicolumn{1}{l|}{0} & \multicolumn{1}{l|}{1} \\
  \hline
 0 & 582 & 14 \\
  \hline
  1 & 6 & 1785 \\
  \hline
  \end{tabular}  } & \multicolumn{6}{c|}{  \begin{tabular}[t]{| c | c | c |}
  \firsthline
  \multicolumn{1}{|l|}{} & \multicolumn{1}{l|}{0} & \multicolumn{1}{l|}{1} \\
  \hline
 0 & 383 & 168 \\
  \hline
  1 & 126 & 1710 \\
  \hline
  \end{tabular}    } \\ \hline

\hline
\end{tabular}
\caption{\small{Confusion Matrices of Supervised Learning Based Classifiers for Sentiment Classification.}}
\label{table:SentimentConfusionMatrix}
\end{table*} \normalsize


\small
\begin{table*}
\centering
\begin{tabular}{|ccccccccccccccc} \cline{4-15}
\multicolumn{3}{c|}{} & \multicolumn{6}{|c|}{Unbalanced Dataset} & \multicolumn{6}{c|}{Balanced Dataset} \\ \cline{1-15}

\multicolumn{1}{|c|}{Model} & \multicolumn{2}{c|}{Class} & \multicolumn{2}{c|}{Precision} & \multicolumn{2}{c|}{Recall} & \multicolumn{2}{c|}{F1-Score} & \multicolumn{2}{c|}{Precision} & \multicolumn{2}{c|}{Recall} & \multicolumn{2}{c|}{F1-Score} \\ \hline

\multirow{2}{*}{SVM} & \multicolumn{2}{|c|}{Neg} & \multicolumn{2}{c|}{0.60} & \multicolumn{2}{c|}{0.56} & \multicolumn{2}{c|}{0.58} & \multicolumn{2}{c|}{0.55} & \multicolumn{2}{c|}{0.52} & \multicolumn{2}{c|}{0.53} \\

 & \multicolumn{2}{|c|}{Pos} & \multicolumn{2}{c|}{0.72} & \multicolumn{2}{c|}{0.75} & \multicolumn{2}{c|}{0.74} & \multicolumn{2}{c|}{0.71} & \multicolumn{2}{c|}{0.74} & \multicolumn{2}{c|}{0.72} \\ \cline{1-15}

\multirow{2}{*}{\begin{tabular}{@{}c@{}}Na\"ive Bayes\\(Bernoulli)\end{tabular}} & \multicolumn{2}{|c|}{Neg} & \multicolumn{2}{c|}{0.51} & \multicolumn{2}{c|}{0.81} & \multicolumn{2}{c|}{0.62} & \multicolumn{2}{c|}{0.51} & \multicolumn{2}{c|}{0.74} & \multicolumn{2}{c|}{0.60} \\ 

 & \multicolumn{2}{|c|}{Pos} & \multicolumn{2}{c|}{0.79} & \multicolumn{2}{c|}{0.48} & \multicolumn{2}{c|}{0.60} & \multicolumn{2}{c|}{0.78} & \multicolumn{2}{c|}{0.56} & \multicolumn{2}{c|}{0.65} \\ \cline{1-15}

\multirow{2}{*}{\begin{tabular}{@{}c@{}}Logistic\\Regression\end{tabular}} & \multicolumn{2}{|c|}{Neg} & \multicolumn{2}{c|}{0.59} & \multicolumn{2}{c|}{0.56} & \multicolumn{2}{c|}{0.58} & \multicolumn{2}{c|}{0.55} & \multicolumn{2}{c|}{0.52} & \multicolumn{2}{c|}{0.53} \\ 

 & \multicolumn{2}{|c|}{Pos} & \multicolumn{2}{c|}{0.72} & \multicolumn{2}{c|}{0.75} & \multicolumn{2}{c|}{0.73} & \multicolumn{2}{c|}{0.71} & \multicolumn{2}{c|}{0.73} & \multicolumn{2}{c|}{0.72} \\ 

\hline
\end{tabular}
\caption{\small{Perfomance of Supervised Learning Based Classifiers for Sentiment Classification.}}
\label{table:SentimentPerformance}
\end{table*} \normalsize

\subsection{Text Classification Methods}

After the initial pre-processing steps described in section \ref{datasection}, we used simple feature extraction procedures to test our models. For the supervised models, we used a bag-of-words approach for feature extraction using Count Vector (occurrences of tokens in each tweet) from scikit-learn to transform words into numerical features. The topics (vaccine and stock market) were also converted into numbers by dummy coding. For the unsupervised models, we used TF-IDF to transform words into numerical features. Additional feature engineering steps were used to improve performance of the algorithms. Our results indicate that our algorithms perform reasonably both in supervised and unsupervised learning, but further improvements are needed. We will use data augmentation to try to improve the performance our models.

\subsubsection{Motivation for using ML}
Our motivation for using supervised and unsupervised learning to classify the topics and sentiments textual distributions was not the popular goal of improving classification accuracy. We achieved strong results for our baseline supervised classification models, as anticipated. Our interest in using these methods was to study the behavior of TDD under conditions such as classification with and without keywords (top frequency Unigram)  and with and without balanced (more items from one class than the other) sentiment datasets. We observed that the removal of one high frequency keyword from the TDD significantly decreased the performance of all the models, indicating the high sensitivity of such models to the top high frequency words, especially if they are unique to each class, as shown in tables \ref{table:TopicConfusionMatrix} and \ref{table:TopicPerformance}.    

\subsubsection{Supervised Learning}
Raw Text data cannot be fed directly to the algorithms themselves as most of the models expect numerical feature vectors with a fixed size rather than the raw text documents with variable length. In order to address this, we used a bag-of-words approach for feature extraction using Count Vector (occurrences of tokens in each tweet) from scikit-learn to extract the features. Once the features are extracted, we feed it to the models we experimented: Logistic Regression, SVM and Na\"ive Bayes models. 

\paragraph{Labels for market and vaccine texts:} Once the data were cleaned and processed using standard NLP preprocessing methods, the \emph{Text} variable was extracted and cleaned, and topic labels (\emph{M} for 2,897 market tweets and \emph{V} for 9,036 vaccine tweets) were added. The labelled text variables from the market-tweets and vaccine-tweets were then combined and their order was randomized. This constituted the main dataset with nearly 12000 tweets for supervised learning. Given our interest in understanding the behaviour of TDD, we found it interesting to repeat the process with a reduced dataset, where we removed the word ``vaccine" from the vaccine dataset and the word ``market" from the market dataset and repeated the process above. We used a 80:20 split to use 9546 tweets for training and tested on 2387 tweets. We used three supervised classification methods, \texttt{SVM, Na\"ive Bayes (Bernoulli) and Logistic Regression}, for each of the above and the resulting confusion matrix and evaluation metrics are provided in  Tables \ref{table:TopicConfusionMatrix} and \ref{table:TopicPerformance} respectively. 

\paragraph{Sentiment classification process:} We also added sentiment labels for positive and negative tweets. The tweets were assigned a sentiment score using the \texttt{SentimentR} package, and the default \texttt{Jockers} dictionary. All tweets with scores greater than 0, were classified as positive tweets and all tweets with scores less than 0 were classified as negative tweets. Neutral tweets with a sentiment score of 0 were ignored, to create a positive - negative labeled primary dataset of 9846 labelled tweets, with 5876 positive and 3970 negative tweets. Since we are interested in studying, understanding and aligning TDD, we found it necessary to repeat the process with a balanced dataset, where we first took an equal number of tweets from each of the datasets (2897 each, from market and vaccine datasets), and then we repeated the process above to exclude neutral tweets leading to a balanced sentiment class dataset of 4849 tweets (by deletion of odd number of neutral tweets). We used a 80:20 split to use 3879 tweets for training and tested on 970 tweets. We used three supervised classification methods, \texttt{SVM, Na\"ive Bayes (Bernoulli) and Logistic Regression}, for each of the above and the resulting confusion matrix and evaluation metrics are provided in Tables \ref{table:SentimentConfusionMatrix} and \ref{table:SentimentPerformance} respectively. 

\begin{small} 
\noindent \textbf{Examples of misclassified tweets:}

\noindent Vaccine tweets misclassified as market tweets:\\
\end{small}
\begin{tiny}
\texttt{
my arm sore from my covid vaccine\\ 
------------------------\\
friends who have recovered from covid and gotten the vaccine  what were your postshot symptoms\\
}
\end{tiny}

\begin{small}
\noindent Misclassified sentiment tweets - negative tweets classified as positive:\\ 
\end{small}
\begin{tiny}
\texttt{
the stock market is bleeding i am bleeding lol\\
------------------------\\
northkhalea little things like walks to the local shop or market is something i definitely overlooked the importance of precovid but i m glad to hear you re carving out your own little corner\\
}
\end{tiny}

\paragraph{SVM:} This model maps training examples to points in a high-dimensional feature space, in order to maximize the
width of the distance between the categories. A hyperplane is built, so that new samples (e. g. the test set) can be classified. The performance achieved with this classifier is reasonable, since we used a very simple linear classification to perform the task as a baseline. It wrongly classified the sentiment for the two examples listed, but also misclassified the topic classes for the examples provided. It is probably put too close to the vaccine topic, because of words as ``bleeding'' and ``precovid''.

\paragraph{Na\"ive Bayes (Bernoulli):} This model is a simple probabilistic classifier built upon Bayes' theorem and the assumption that features are independent. The performance of our model was surprisingly the best in the Topic Classification task, which can be due to the fact that we used linear classifiers in SVM and Logistic Regression 
and that our feature extractor was based on word frequencies. In the Sentiment Classification task, the performance was better in the negative class, but worse in the positive class. While it also misclassified the sentiment class of the examples, it did correctly classify a tweet in which the words "stock" and "market" are present. That illustrates the better performance achieved by the Na\"ive Bayes, since these words increase the probability that it belonged to the class \textit{Market}.

\paragraph{Logistic Regression:} As with the SVM model, the logistic regression is also a simple linear classifier. The predictor is a linear equation that is mapped into a binary classification by a logistic link function. As expected, the performance of our logistic regression was very similar to that of the SVM. The two examples listed were misclassified by our logistic regression model too. They show that probably the model is associating the word ``vaccine'' with a positive sentiment, but it is not giving the proper weights to negative words, such as ``sore'', or to sequences, such as ``postshot symptoms''.


\subsubsection{Unsupervised Learning}

For textual classification based on unsupervised learning, we decided to explore two clustering methods: (i) Hierarchical Agglomerative Clustering (HAC) and (ii) K-Means Clustering (KMC). After a first round of evaluation, we tried to combine the most successful method with Independent Component Analysis (ICA). We present both of them and explore the initial results we obtained running them against our labeled data.

\paragraph{Hierarchical Agglomerative Clustering} -- This unsupervised method groups together observations whose features are similar. For the sake of illustration, we show how 100 occurrences of our data would be hierarchically clustered in Figure \ref{fig:HACtweets}.


After recursively and hierarchically merging pairs of clusters increasing the linkage distance as less as possible, clusters are naturally formed. We chose 2 clusters, since we are interested in getting as close as possible to the annotated topics. After training on 9546 tweets, the algorithm indicated two unbalanced classes, overlapping in 73\% with our manually annotated classes.

\begin{table}
\small
\centering
\begin{tabular}{c c c c c}
\hline 
 &precision&recall&f1-score&support\\
\hline \hline
class 0 &0.76&0.95&0.84&7253\\
\hline
class 1 &0.17&0.03&0.03&2293\\
\hline
accuracy & & &0.73&9546\\
\hline
macro avg &0.46&0.49&0.45&9546\\
\hline
weighted avg &0.61&0.73&0.65&9546\\
\hline
\end{tabular}
\caption{\small{Classification Report on HAC}}
\label{table:HAC}
\end{table} \normalsize

\paragraph{K-Means Clustering} -- This unsupervised method also groups together observations whose features are similar, but the procedure does not rely on recursively merging pairs, but rather creating a mean prototype (cluster center or centroid) and clustering the others according to the distance to the centroid. For our test case, we set up two clusters aimed at overlapping with the two topics that we had manually annotated. The results are summarized in Table \ref{table:KMC}.

\begin{table}
\small
\centering
\begin{tabular}{c c c c c}
\hline 
 &precision&recall&f1-score&support\\
\hline \hline
class 0 &0.70&0.71&0.70&7253\\
\hline
class 1 &0.04&0.04&0.04&2293\\
\hline
accuracy & & &0.55&9546\\
\hline
macro avg &0.37&0.38&0.37&9546\\
\hline
weighted avg &0.54&0.55&0.55&9546\\
\hline
\end{tabular}
\caption{\small{Classification Report on KMC}}
\label{table:KMC}
\end{table} \normalsize

\paragraph{Independent Component Analysis} --  This unsupervised method is a generative model to reveal hidden factors that underlie a set of features. Often some subcomponents of the features are statistically independent from each other. The goal is to raise components that are maximally independent. We used this method in combination with the HAC to try to get an improvement in the performance of our algorithm. As summarized in Table \ref{table:HACwithICA}, the accuracy improved by almost 1\% only adding ICA and holding everything else constant.

\begin{table}
\small
\centering
\begin{tabular}{c c c c c}
\hline 
 &precision&recall&f1-score&support\\
\hline \hline
class 0 &0.76&0.96&0.85&7253\\
\hline
class 1 &0.19&0.03&0.05&2293\\
\hline
accuracy & & &0.74&9546\\
\hline
macro avg &0.47&0.50&0.45&9546\\
\hline
weighted avg &0.62&0.74&0.66&9546\\
\hline
\end{tabular}
\caption{\small{Classification Report on HAC and ICA}}
\label{table:HACwithICA}
\end{table} \normalsize

\section{Text generation}
Text generation has been been addressed since the early 1950's and has since evolved into a science with an array of sophisticated methods to address a broad range of NLP challenges \citep{klein1963syntactic}. State of the art transformer driven models today include BERT, GPT, XLNet and ELMo \citep{topal2021exploring, ethayarajh2019contextual}. There is strong and growing interest in the field of text generation and natural language generation at multiple levels, and many innovative approaches including adversarial strategies are being attempted to improve different dimensions of text generation \citep{zhang2017adversarial}. We use three methods to generate text for testing the alignment of resulting distributions. We provide a brief description of each method, along with preliminary illustrative output are provided below. Though we use direct probabilistic and RNN-LSTM as the first two methods to generate text, extant research identifies problems with previous methods such as RNN-LSTM and posits the advantages of using GPT. However GPT, BERT, ELMo, GloVe (Global Vectors for Word Representation) and FastText each have their own strengths in text generation \citep{ethayarajh2019contextual}. Based on our research goals, we found it most relevant to use fine tuned GPT-2 models as our third method for text generation.    

\subsubsection{Direct Probabilistic Text Generation}
In the first method, we extracted the text variable (TextMV) and fed it into a Jupyter notebook environment with NLTK and Pandas. TextMV was then pre-processed and cleaned using NLTK, and the words in TextMV were tokenized, and converted to a dataframe with 2991 unique tokens. Probabilities were then calculated based on the relative position of each token, and text was generated based on seed words, which needed to be any of the 2991 tokens. Illustrative output showing the text generated using this method and keyword  ``vaccine'' is posted below, the length of each output was limited to 40 words:    

\begin{tiny}
\texttt{
covid vaccine powers will need to let me take some extra doses separate issues and getting my sleeves were quite fatigued and puppet government might be ordered to a vaccine centers not about other than we go to do we\\ 
------------------------\\
covid b had our bio to know how to the age limit to see it sounds like service orgs with jayson werth during a wentz sb jersey my boys love the job i would you want to a free market\\
------------------------\\
covid vaccine doses were slated to have to live in our rights as other verticals to protect ourselves and bring to suffer the ads just chill out andrewsiffert ayayrongreen from now the vaccine doesnt get another leftard who prefer hot\\ 
------------------------\\
covid vaccine amp they opening and more to be good thing as bonus the sox traded company if they go to meet rights fee demands can spend any restaurants you might make sure if it rachidrouis free market posts timpc9213\\ 
------------------------\\
covid vaccine but typical for the next agenda dupped usefulidiots last wednesday at the county collecting ring was able to drop for the market backwards our entire career thanks is hesitant to these would be easy for buyers right to\\ 
------------------------\\
covid vaccine passport and speculative picks even close to do not sure thanks to have to a feeling blessed to travel and walls of everything teapainusa destroy people die because eg there was growing up rolling over the quantity of\\ 
------------------------\\
covid vaccine records bc why cant see our local stations facebook page are the higher rate swap market until then why we just a good to agbanker for on market for my thoughts saviroman matthaneysf around farmers market cap of }

\end{tiny}

\subsubsection{LSTM Text Generation} 
With the second method, we extracted the text variable(TextMV) and fed it into a Jupyter notebook environment with TensorFlow, Scikit-Learn and Numpy. TextMV was then pre-processed and cleaned to create a raw text file, and then we built a LSTM model with 30 epochs, and repeated the exercise with 7 epochs. The output displayed below is based the model generated with 7 epochs, initiated by four seed entries, and limited to under 80 words:

\begin{tiny}
\texttt{
had nothing to do with developing the covid vaccine i suppose next hell be credited with inventing the wheel vaccine hunting is like amiibo hunting so after collecting ring was she supposed to continue pricing tomato thy market proposal funny dieeeeee supposedly i had my st vaccination shot today but am not sure if it actually happened because didnt watch the needle puncture the skin amp didnt feel a single thing vaccine u f nolau cufe f due to texas weather uncertainty florida will boom even faster now better weather no state taxes too the fl housing market shortage will be see to be able to apply get a vaccine passport to travel and a lot of our lot of apply vaccine amp abuvs just just like like of the vaccine amp yeah the market is going to make a copy market i am to do you will have a lot of it s market and not the just just just just like of the vaccine thats its my arm vaccine and i m just just like\\
------------------------\\
the question is are we expected to have a another decade bull run given shiller s pe ratio averages x super interesting call out andrewsiffert ayayrongreen from a market share perspective who are the top carriers in the region i think real estate and the stock market are the two best use cases for blockchain that are hardly being utilized yet it s funny how capitalism s whole thing is no monopolies the power outages in texas and in louisiana are due to these companies owning the energy market and getting away with murder can you recommend anyone for this technician covid vaccine support at a year yeah that i m to get the vaccine and i am to do you are been to do i m not not to get the vaccine shot to travel and i m to be a appointment i am to get a vaccine passport to get a vaccine passport to travel and a lot of our bio for apply vaccine passport because the market for this abuvs abuvs i had my \\ 
------------------------\\
no where is the market for jackie bradley now redsox should not pay for him because they are in rebuilding mode but he is a winner and so well liked in boston hmmm royalcaribbean requiring vaccine i wasnt planning to get one but you guys changed my mind get my second vaccine shot today and im kinda nervous u f f idk why we need to develop the market for some products here too i mean made in china to spice things up took the covid vaccine and im just left with covid u f a miamillerx market in rocky river and was the flea market is the line of the market just just just just just just just on the second vaccine i am to get the vaccine thats the arm is to get a vaccine passport to get a vaccine passport to travel and a lot of our bio for apply vaccine passport because the market for my second vaccine and be no little than no arm you are to do and i m not \\ 
------------------------\\
travel is the same as needing other well known vaccinations for international travel if i have to get it to travel iiiii will but i also will not act like its the same as other required vacs and make ppl feel bad if you dont wanna get the vaccine thats fine but if your gunna try to convince people not to get it ur a weirdoooooo alexberenson didnt know what vaers is until getting my vaccine and being told about it signed up to report side effects that are nominally nonexistent this infers many more peeps on a percentage basis to travel a lot of this market amp open abuvs a lot of getting my second vaccine shot and im a good thing if you dont get an appointment i got my second vaccine and the power year in the last world and are market and like like their power grid in our bio to apply their covid vaccine support vaccine im abuvs but you have a appointment to get the covid vaccine support and im abuvs }

\end{tiny}

\subsubsection{GPT-2 Text Generation} 
With the third method, we used Azure to fine-tune GPT2 nuermous times. Initially we used the text variable(TextMV) and generated text with fine-tuned GPT-2. Based on the relatively greater superiority of readability and coherence of text generated with GPT-2 as compared to the first two methods, we chose to generate the final textual datasets to test for distributional alignment using GPT-2 on Azure. We fine tuned GPT-2 models by topic, Vaccine and Market, and by sentiment, positive and negative, by fine-tuning GPT-2 repeatedly with Vaccine, Market, Positive sentiment and negative sentiment tweets, respectively. The output from GPT-2 for these categories is displayed below. 
\paragraph{GPT-2 text generation for Vaccine topic} These generated texts were mostly on topic, with a few stray items. Some items were creatively structured by the finetuned model: 

\begin{tiny} 
\texttt{
I’m getting my COVID vaccine today, so check back for my review on that too. I had some tough decisions to make along the way, and having those decisions be that I’m not going to get tested for Cov\\ 
------------------------\\
@Burn\_the\_ships @Mack3211 Yeah, I get it. But the vaccine passport is just a way for the government to collect and sell your information, basically.
@mack\_riley @AriFleischer Imagine\\ 
------------------------\\
I got my second dose of the vaccine today and I feel like my arm is about to fall off <U+0001F97A> + I haven’t put any weight on my right arm <U+\\ 
------------------------\\
@annabkrr Keep wearing that mask, Nana. I got my second shot and still wear two masks and a face visor. The vaccine works for the original,  wild\\ 
------------------------\\
A new study shows that not only has the COVID vaccine made women infertile ``I don’t get why a vaccine passport is bad or a breeding ground for West African recurrences, but that doesn’t mean\\ 
------------------------\\
I got the first dose of the Pfizer vaccine today and I’m feeling the side effects pretty bad. I’ve been doing a lot of reading online about the long term effects of the vaccine and how to mitigate them.\\ 
------------------------\\
Founded in 1859, Milledie’s Ice Cream parlors are the best in town. There are a handful of “second chances”, but most are roaring success. 2nd Milledie’}

\end{tiny}

\paragraph{GPT-2 text generation for Market topic} 
It is interesting as to how the model makes an effort to mimic tweets even at the character level, however, it does appear to miss some context:

\begin{tiny}
\texttt{
Okay, the ``let the market sort it out" option seems to be the better one. Buyers should be able to settle for substandard products knowing that even if they hammer out an insane price, it’ll be far lessening than if the product had been offered at that price.\\ 
------------------------\\
Maybe it’s the @Browns saying “trust but verify” when selling high. If Gordon Hayward goes WRB, this could be a nice driver to help you score. If not, it’s trade chips.\\ 
------------------------\\
@myfirstpassengers Yeah, I guess that's why they put the stock in the market! Makes sense to me.\\ 
------------------------\\
As 2020 likely bookends a distinct era which we cannot predict with precision “surely a reasonable 5\% error margin.\\ 
------------------------\\
@SalariesAreStolen Again... Affirmative action pays less than market rate per hour.. wait a minute... what? \\ 
------------------------\\
@DanielGullotta @Criterion Thief would be a must. Unfortunately,  I think Charade is off the market now.\\ 
------------------------\\
@cmarchena @RudrakshPande19K But the stock market and all those options are where we are now with regards to credit cards and other types of products. I'm not sure if there's a readymade plan for those. But I sure as 7abuvs
}

\end{tiny}

\paragraph{GPT-2 text generation for positive sentiment} Most of the text items generated positive sentiment, but there were a high amount of matched phrases between the generated and seed text: 

\begin{tiny}
\texttt{
A HUGE shoutout to @DallasFireRes\_q  working the @KBHCCDallas vaccine station 24 today!!  Your kindness \& professionalism made the \#COVIDVaccination experience\\ 
------------------------\\
yoo just gave me an update on the second vaccine location very grateful\\ 
------------------------\\
I help two older Americans get vaccine appointments today which is almost as impressive as helping someone acquire the new Xbox.\\ 
------------------------\\
I laugh when I tell people Im not taking the Vaccine and they say  but your already vaccinated yes because I was a child and I did what my mother told me to doU0001F9\\ 
------------------------\\
athena89152 iamryanjtrump ja11eo24 Excited to see how these two approaches to cancer compare  I know I wont be getting the vaccine but I think its important\\ 
------------------------\\
@gforce\_bg Yes. Plus everyone (well 99\%) are happy to be there, so we're happy to be there with them. It's rewarding to be able to calm someone who is\\ 
------------------------\\
LurkingFinn The vaccine isnt 100 protected But it is much safer than when I was a kid hoping that vaccine with the vaccinell protect me and others as well U0001F643\\ 
------------------------\\
TimKilleen ChaChaCostaMD  Ive been enjoying s normal life since the beginning of thisworking traveling celebrating Christmas Thanksgiving with multiple households in multiple cities going out to eat shoppingetc\\ 
------------------------\\
@ProfMattFox AZ seems like a particularly lower-quality vaccine compared to even technically-similar J\&J, but yes we of course need to watch that.\\ 
------------------------\\
selenarosemary and I just spent literally 24 hours on the couch streaming movies and recovering from vaccine 2 feeling great now very much to the delight of our dog Chewie U0001F415
}

\end{tiny}

\paragraph{GPT-2 text generation for negative sentiment} Most of the items generated contained negative words, but some of them did not have negative meanings in spite of the use of negative words: 

\begin{tiny}
\texttt{
khuwig1 ohiodata The vaccine is going to make people sick The actual virus is going to kill people worldwide\\ 
------------------------\\
Had Covid last year and was very very sick Thought I may die Took 7 months for my lungs to recover\\ 
------------------------\\
Kierz10 zeynepyenisey I agree that Covid is a flu level risk for a healthy 26 year old That doesnt contradict what I wrote Getting the flu is more dangerous than\\ 
------------------------\\
NayriiTime People have been traveling the entire Pandemic without a vaccine Theyre ridiculous with these conspiracy theories\\ ------------------------\\
my grandma crazy af talking bout she getting the vaccine\\ 
------------------------\\
@SenSanders Bernie. If my car has a defect \& injures me I can sue the maker. If my vaccine shot injures me, I can’t sue the maker.\\ 
------------------------\\
Meteor Shower Nearby houses on lockdown due to an unrelated and as of yet unconfirmed incident. All available shelters full. Gov DeSantis only talks about the vaccine. No restrictions put in\\ 
------------------------\\
I should have known that this vaccination roll out would be a disaster. Grandmas 2nd vaccine is due today and nobody has contacted us about the 2nd shot and the Escondido location she got\\ 
------------------------\\
Howdyhaylee Its insanity Unfortunately theres no vaccine for that line of thinking\\ 
------------------------\\
Suddenly all the antivax conspiracy theorists are blaming the vaccine for everything from acne to acne- the ravages of time. Me neither. I grew up in a house without a vaccine, and
}

\end{tiny}

\section{Development of KL Textual Distributions Contrasts}

Consider our original distribution of interest $Vo$ whose nature we are interested in replicating as a machine generated distribution $Vg$. Having generated $Vg$ through the process described above, we are now interested in applying KL-Divergence to study the alignment of the machine generated distribution $Vg$ with the original distribution $Vo$:
\begin{align}
\displaystyle\
KL(Vo||Vg) &=  \int_{-\infty}^{\infty} Vo(x) [\log\frac{Vo(x)}{Vg(x)}]dx
\end{align}

Which in our case, for discrete word count based distributions, leads to:
\begin{align} \label{eq02}
\displaystyle\
KL(Vo||Vg) &= \sum_{x \in X} Vo(x)[\log\frac{Vo(x)}{Vg(x)}]
\end{align}

We are interested in the relative entropy of the generated TDD compared to the original TDD, and therefore we do not attempt to apply the symmetric form of KLD. 

The TDD alignment validation process begins with a unique approach to generating TDD aligned by topic and by sentiment, which is very efficient for short texts such as tweets, and can be applied to longer corpora with minimal adjustments. Consider the original text of the vaccine tweets and market tweets, $TwV_{Original}$ and $TwM_{Original}$, respectively, which are processed into  words $W$ indexed in token style as $j$ and decreasingly sorted as unigrams with frequencies $\alpha$: 
\begin{align}
\displaystyle\
    TwV_{Original} \rightarrow  W_{j\alpha} V_{Original} \\
    TwM_{Original} \rightarrow  W_{j\alpha} M_{Original}
\end{align}
A similar process applied to the generated topical distribution texts will lead us to: 
\begin{align}
    TwV_{Generated} \rightarrow  W_{j'\alpha} V_{Generated} \\
    TwM_{Generated} \rightarrow  W_{j'_\alpha} M_{Generated}
\end{align}
However, when assigning indices for words from the generated textual distribution, the generated word indices are matched to the original word indices: for example, a word ``price'' in $TwTopic_{Generated}$ will have same index $j'$ value assignment as the index $j$ value assignment in $TwTopic_{Original}$. Furthermore, it is important to account for unique words in $TwTopic_{Original}$, the index values of which are included for $j'$ in $TwTopic_{Generated}$ with $\alpha = 0$. Then the $i$ number of unique words in $TwTopic_{Generated}$ are then appended to index $j$ in $TwTopic_{Original}$ with $\alpha = 0$, such that the final index $(j + i)$ of $TwTopic_{Original}$ will be a perfect match with $j'$ of $TwTopic_{Generated}$. It is possible that in some cases such $i = 0$, implying that there are no words in $TwTopic_{Generated}$ which are not already included in $TwTopic_{Original}$. Some data scientists prefer to use $(j - i_o -i_g)$, implying a reduction of unique words from both $TwTopic_{Original} = i_o$ and $TwTopic_{Generated} = i_g$, to identify and subset words common to both data. We chose to start with the $(j + i)$ approach, and then retain the option to select a predetermined number of common words with highest frequencies at the point of calculating the KLD values. Therefore, after applying the algorithmic index matching process between $TwTopic_{Original}$ and $TwTopic_{Generated}$, the generalization of the equations above are rewritten as: 
\begin{align}
\displaystyle\
    \label{eq07} TwTopic_{Original} \rightarrow  W_{(j+i)\alpha} TwTopic_{Original} \\
    \label{eq08} TwTopic_{Generated} \rightarrow  W_{(j+i)\alpha} TwTopic_{Generated}
\end{align}
leading to: 
\begin{align}
\displaystyle\
    TwV_{Original} \rightarrow  W_{(j+i)\alpha} V_{Original} \\
    TwM_{Original} \rightarrow  W_{(j+i)\alpha} M_{Original}
\end{align}
A similar process applied to the generated topical distribution texts will lead us to: 
\begin{align}
    TwV_{Generated} \rightarrow  W_{(j+i)\alpha} V_{Generated} \\
    TwM_{Generated} \rightarrow  W_{(j+i)\alpha} M_{Generated}
\end{align}

So also, we classify TDD alignment based on sentiment, wherein the original text of the vaccine tweets and market tweets are combined and classified as being positive or negative (neutral tweets are ignored), $TwPos_{Original}$ and $TwNeg_{Original}$, respectively, which are processed into  words $W$ indexed in token-style as $j$ and decreasingly sorted as unigrams with frequencies $\alpha$. We start with the generalization for sentiment:
\begin{align}
    TwSenti_{Original} \rightarrow W_{(j+i)\alpha} Senti_{Original} \\
    TwSenti_{Generated} \rightarrow W_{(j+i)\alpha} Senti_{Generated} 
\end{align}\\
Leading to:
\begin{align}
    TwPos_{Original} \rightarrow  W_{(j+i)\alpha} Pos_{Original} \\
    TwNeg_{Original} \rightarrow  W_{(j+i)\alpha} Neg_{Original}
\end{align}
A similar process applied to the generated sentiment distribution texts will lead us to 
\begin{align}
    TwPos_{Generated} \rightarrow  W_{(j+i)\alpha} Pos_{Generated} \\
    \label{eq18} TwNeg_{Generated} \rightarrow  W_{(j+i)\alpha} Neg_{Generated}
\end{align}

The frequencies ``$\alpha$'' are then normalized using a Softmax function within $TwTopic_{Original}$  \&  $TwTopic_{Generated}$ each \\ 

and within \\ 
$TwSenti_{Original}$ \& $ TwSenti_{Generated} $ each.\\

The general multi-class Softmax function for a single label classification is given by
\begin{align}
\displaystyle\
\sigma(z_i) = \frac{e^{z_{i}}}{\sum_{j=1}^K e^{z_{j}}} \ \ \ for\ i=1,2,\dots,K
\end{align} \\

For our purposes, this is simplified to: 
\begin{align}
\displaystyle\
\sigma(\alpha_{[j+i]}) = \frac{e^{\alpha_{[j+i]}}}{\sum_{h=1}^L e^{\alpha_{h}}} \ \ \ for\ h=1,2,\dots,L
\end{align} \\

Applying the Softmax to the $\alpha$ (frequency) vector of each of the distributions allows us to use \texttt{KLD} meaningfully to test the alignment of textual distributions because it enables an index matched and proportionate contrast, i.e. an index matched distance summary, and the use of the Softmax function ensures that the size of the generated textual corpora does not matter, subject to a heuristic and contextual minimum size. Now we are able to contrast the distributions using \texttt{KLD} by applying equations \ref{eq02}, \ref{eq07}, \ref{eq08} and \ref{eq18}:  
\begin{align}
\displaystyle\
\textit{For all [j+i] = x, let: } [j+i] \in X 
\end{align} 
\paragraph{KL Textual Distributions Contrasts (KL-TDC)}
Then for all $\alpha_{[j+i]} = \alpha_{x}$, we can develop a general application of our \texttt{KLD} measure between any two distributions $V_\pi$ and $V_\phi$, where $V_\phi$ is the standard distribution and $V_\pi$ is the distribution we seek to evaluate for relative entropy:
\begin{align} \label{eq22}
\displaystyle\
KL(V_\phi\alpha||V_\pi\alpha) &= \sum_{x \in X} V_\phi\alpha(x)[\log\frac{V_\phi\alpha(x)}{V_\pi\alpha(x)}]
\end{align} \\
\texttt{KL-TDC} thus obtained is a contextual measure: the metric obtained by applying \texttt{KL-TDC} will need to be compared to another ``baseline'' \texttt{KL-TDC} metric. Such a baseline metric can be obtained in a number of ways, subject to the objectives, nature of the TDD scenario and the availability of additional naturally occurring $Text_Original+$ data that can be compared to the $Text_Original$ data. If such additional naturally occurring $Text_Original+$ data are not available, then a random sampling process can be used to draw samples from $Text_Original$ data, and then used for comparison. The method logical process aspects are elaborated under the Experimental Results section below.

Applying the \texttt{KL-TDC} equation \ref{eq22} to our scenario for comparing original ($ To = W_{(j+i)\alpha} TwTopic_{Original} $) and generated ($Tg = W_{(j+i)\alpha} TwTopic_{Generated} $) topic distributions we have:  
\begin{align}
\displaystyle\
KL(To\alpha||Tg\alpha) &= \sum_{x \in X} To\alpha(x)[\log\frac{To\alpha(x)}{Tg\alpha(x)}]
\end{align}

So also, we extend the \texttt{KL-TDC} equation \ref{eq22} to our scenario for comparing original sentiment ($So = W_{(j+i)\alpha} Sentiment_{Original}$) and generated sentiment ($Sg = W_{(j+i)\alpha} Sentiment_{Generated}$) textual distributions we have:  
\begin{align} 
\displaystyle\
KL(So\alpha||Sg\alpha) &= \sum_{x \in X} So\alpha(x)[\log\frac{So\alpha(x)}{Sg\alpha(x)}]
\end{align} 

\subsection{Applied KL Textual Distributions Contrasts (\texttt{KL-TDC})}

We applied the \texttt{KL-TDC} metric to the scenarios listed below and identified the measure to which different TDD were aligned with each other. These five scenarios represent the completion of the TDD generation process, and then we present \texttt{KL-TDC} metrics for these scenarios under the experimental results section following the description of the scenarios. 

\subsubsection{Vaccine}
We finetuned GPT-2 on Azure with $9,036$ $TwVac_{Original}$ vaccine tweets, and generated text $TwVac_{Generated}$ with the vaccine-finetuned GPT-2 model. $TwVac_{Generated}$ was then fed into our Unigram algorithm, and the frequencies, $\alpha$ values, were then normalized with the Softmax function adapted to a simple count scenario. A similar process was repeated with $TwVac_{Original}$ and the two resulting probability vectors based on the 100 top unigrams from $TwVac_{Original}$ were fed into \texttt{KL-TDC} to obtain the TDD alignment score.

\subsubsection{Market}
We finetuned GPT-2 on Azure with $2,897$ $TwMkt_{Original}$ market tweets, and generated text $TwMkt_{Generated}$ with the market-finetuned GPT-2 model. $TwMkt_{Generated}$ was then fed into our Unigram algorithm, and the frequencies, $\alpha$ values, were then normalized with the Softmax function adapted to a simple count scenario. A similar process was repeated with $TwMkt_{Original}$ and the two resulting probability vectors based on the 100 top unigrams from $TwMkt_{Original}$ were fed into \texttt{KL-TDC} to obtain the TDD alignment score.

\subsubsection{Positive}
In this scenario, we moved from topic parameters to sentiment parameters: We finetuned GPT-2 with positive sentiment tweets and generated a positive sentiment based textual data distribution. Given the challenges associated with neutral and near-neutral sentiment tweets, we excluded all tweets with a $Sentiment Score < 0.4$ in our positive tweets corpus $TwPos_{Original}$.   We finetuned GPT-2 on Azure with $883$ $TwPos_{Original}$ positive tweets, and generated text $TwPos_{Generated}$ with the positive-sentiment-finetuned GPT-2 model. $TwPos_{Generated}$ was then fed into our Unigram algorithm, and the frequencies, $\alpha$ values, were then normalized with the Softmax function adapted to a simple count scenario. A similar process was repeated with $TwPos_{Original}$ and \texttt{KL-TDC} was applied to the two resulting probability vectors based on the 100 top Unigrams from $TwPos_{Original}$, to obtain the positive sentiment TDD alignment score.

\subsubsection{Negative}
For this scenario, we repeat the process used for generating $TwPos_{Generated}$ above: We finetuned GPT-2 with negative sentiment tweets and generated a negative sentiment based textual data distribution. Applying the same principle as for $TwPos_{Generated}$ above, we excluded all tweets with a $Sentiment Score > -0.4$ in our negative tweets corpus $TwNeg_{Original}$.   We finetuned GPT-2 on Azure with $521$ $TwNeg_{Original}$ negative tweets, and generated text $TwNeg_{Generated}$ with the negative-sentiment-finetuned GPT-2 model. $TwNeg_{Generated}$ was then fed into our Unigram algorithm, and the frequencies, $\alpha$ values, were then normalized with the Softmax function, as in above scenarios. A similar process was repeated with $TwNeg_{Original}$ and \texttt{KL-TDC} was applied to the two resulting probability vectors based on the 100 top Unigrams from $TwNeg_{Original}$, to obtain the negative sentiment TDD alignment score.



\subsection{Experimental Results}
In our experimental analysis of the scenarios described above, we identified potential baseline scores to make relative sense of the \texttt{KL-TDC} metric, since  is a \texttt{KL-TDC} contextual measure that needs to be compared to a baseline \texttt{KL-TDC} metric for each scenario. The baselines \texttt{KL-TDC} scores were computed by drawing a random sample of approximately $10\%$ of the total tweets in each distribution.  Table \ref{table:KL} below summarizes the results of the experiments. Overall the generated TDD performed well and did not stray too far away from the original TDD or the baseline distributions. A well aligned distribution will have a low \texttt{KL-TDC} score below $1$, for example the \texttt{KL-TDC}, where the two distributions are exactly identical $P(o)==P(g)$, is given by $KL-TDC(P(o)||P(g)) = 0 $.

\begin{table}[!ht]
\small
\centering
\begin{tabular}{c c c c }
\hline 
 TDD&Generated&Baseline&B:G\\
\hline 
Vaccine  &0.079&0.016&0.195\\
\hline
Market   &0.082&0.047&0.58\\
\hline
Positive &0.058&0.089&1.55\\
\hline
Negative &0.077&0.072&0.94\\
\hline
\end{tabular}
\caption{\small{Experimental \texttt{KL-TDC} results}}
\label{table:KL}
\end{table} \normalsize

The baseline Vaccine distribution turned out to be extremely well aligned with the original distribution, while all generated distributions performed well with \texttt{KL-TDC} $< 0.1$. The $B:G$ ratio is a quick summary of how well the generated distribution compares to the baseline, and value greater than 1, indicates that the generated distribution is better than the baseline reference distribution. For example, the positive generated distributed in particular possessed not only a good intrinsic alignment with the original distribution, but also outperformed the baseline distribution ($B:G = 1.55$).

\section{Discussion}

Developing artificially generated TDD is a broad arena, and poses numerous challenges - we qualify our problem on the basis of prior knowledge of topic and a priori generated sentiment, both categories of which constitute our ``original'' textual distributions. We applied supervised and unsupervised machine learning algorithms on variations of data to develop a deeper understanding of TDD, by repeating topical classification machine learning with a keyword removal based reduced distribution. So also we studied the behavior of sentiment classes with balanced and imbalanced datasets. Our objective was not the intrinsic improvement of ML classification algorithms but an exploration of the behavior of TDD by topic and by sentiment.

We used Twitter data for this study because of the increasing interest in tweets analytics - Twitter data and other short text chat data have been used for a wide range of purposes including the study of COVID-19, public policy, vaccinations and human opinion across disciplines \citep{ali2021public, samuel2020covid, rahman2021socioeconomic, samuel2020feeling, pelaez2021david}. \texttt{KL-TDC} can be directly applied to a broad range of short-text cases, including texts from chats, customer reviews and social media posts. Additional investigation would be required to study the operational nuances associated with applying the \texttt{KL-TDC} measure to longer texts, though we do not see any conceptual problems with an extension of the \texttt{KL-TDC} logic to longer texts.   

In the present study, one of the crucial issues was to develop an effective, parsimonious and extensible method to compare TDD, and we believe that we have made significant progress with the current conceptual and mathematical articulation of \texttt{KL-TDC}. Furthermore, we wanted to implement the entire TDD life-cycle of  acquisition, preparation, classification, parameter specified (topic / sentiment) textual data generation and evaluation of the alignment of such machine generated data with stated generation intent using \texttt{KL-TDC}. We believe that we have achieved a fair degree of success in completing this TDD life-cycle and 
measured the similarity of \texttt{original: artificially generated} data sets. 

\subsection{Implications}
Our study presents interesting implications for practitioners and academics: The \texttt{KL-TDC} measure can serve as a locally objective quantitative measure to evaluate whether the artificially generated data is drawn out of the intended or same (input) distribution or not. 
Therefore \texttt{KL-TDC} can serve as a suitable measure for comparison, to be used to test artificially generated data with natural data, synthetic (mixed) data and other artificially generated data distributions. Practitioners can use this method to ensure: (i) machine generated data posses alignment sufficiency, and (ii) substitute expensive data acquisition or generation methods with more cost effective methods based on a minimum necessary \texttt{KL-TDC} measure for data used.

Academics can use this method and the \texttt{KL-TDC} to generate texts efficiently for classroom and research purposes, and for evaluation of textual data, respectively. Both the methods and the measures used described in this study can be used to extend information facets and behavioral research, for example in behavioral finance \citep{samuel2017information}.  With additional development and extension, we hope that insights from the \texttt{KL-TDC} life-cycle process and measure will mitigate at least partially, the NLP and NLG domain dependence on models with a larger number of parameters trained on a colossal amount of data, such as GPT-3 with 175 billion parameters! \citep{devlin2018bert,  radford2019language, brown2019verbnet, topal2021exploring}.

\subsubsection{Limitations and Weaknesses}
We have identified a few limitations and weaknesses of this study: First, our data is limited by size and scope, and by restricted topical and sentiment contexts. This limitation can be mitigated by expanding the study in the future with a broader array of datasets and empirical studies. Second, even though we have used GPT-2 for our final data generation and validation process, we may eventually need to test with several other suitable external text augmentation models such as \texttt{BERT, GloVe, ELMo, and XLNet} for our artificially generated TDD. Not using external augmentation may overfit the artificially generated textual data to the original data based on topic or sentiment or other textual parameter. 
Third, we have not exhaustively studied existing options for textual data generation and it remains possible that an existing method may already perform what we are attempting or better from a TDD generation perspective - nevertheless, our unique approach to textual data distribution generation and alignment validation will add value to applied frameworks on the subject. Finally, we highlight our focus on distributional text generation, implying that this study had limited interest in the intrinsic item-wise semantics, and sensibility of text generated. 

\section{Future Research and Conclusion}

This study opens a stream of possibilities for TDD generation by conceptual parameters such as topic and sentiment. Other parameters that we intend to investigate in the future include style, temporal ( for example: news) alignment and meaning. We also plan to test our models on additional topics, and explore alternative measures for TDD alignment or similarity. Large language models need to rely on high performance computing (HPC) and this is becoming increasingly viable with efforts to expand access to super-computing and HPC democratization initiatives \citep{samuel2021strategies}. However, HPC hours are expensive and comes with their own operational challenges, along with sustainability issues. Therefore, it is important to develop methods and processes which support TDD generation on personal computers, with sufficient levels of quality - this will be of immense help to practitioners, researchers and for classroom use.

Our goals for this study, which represents phase-2 of our research stream on applied textual analytics, TDD, NLG and meanings in NL, were  to:
(i) Explore the behavior of textual classification models with supervised and unsupervised machine learning methods, 
(ii) Develop a process that supports the alignment of generation of textual distributions by topic and by sentiment,
(iii) Generate three levels of text - random intrinsic topic aligned textual data generation with direct probabilistic models, topic aligned semi-structured data generation with RNNs and and LSTM and structured textual data generation with external textual data augmentation, by topics and by sentiment, with GPT2, and most importantly, what all of the above is leading to, (iv) Development of the KL Textual Distributions Contrasts (\texttt{KL-TDC}) process and metric. We have accomplished all of our goals, and have made a notable contribution to the domain of efficient TDD alignment, generation and validation.  

In doing so, we have successfully demonstrated the merit of our propositions. While it remains possible in the future, that these propositions may be further refined as we improve our conceptual understanding and develop associated metrics and models, it is evident that the ground work for successfully accomplishing this has been laid out. We believe that having demonstrated the entire TDD life-cycle of  acquisition, preparation, classification, parameter specified (topic / sentiment) textual data generation and evaluation of the alignment of such machine generated data with stated generation intent using \texttt{KL-TDC}, future research can now extend this valuable stream of research to improve both the efficiency of distributional text generation, as well the effectiveness with which the qualitative parameters of such machine generated text can be controlled, including the use of alternative methods, for example, to generate tweets with high popularity potential for going viral \citep{garvey2021would}. Given the current trajectory of this research, we anticipate sustainable and useful contributions to the NLP and NLG through the use and further development of \texttt{KL-TDC}.\\ 

\small
\section*{Acknowledgement}
This study was initiated as part of the Artificial Intelligence certification program at Stanford University. We acknowledge the valuable advisory support from faculty and staff at the Stanford Center for Professional Development, and in particular, Dr. Khalid Abbas El-Awady. 
Limitations or errors, if any, are our own. 

\bibliographystyle{acl_natbib}
\bibliography{acl2021}


\end{document}